\RequirePackage{rotating} %bea
\documentclass[acmsmall]{acmart}

\AtBeginDocument{%
  \providecommand\BibTeX{{%
    \normalfont B\kern-0.5em{\scshape i\kern-0.25em b}\kern-0.8em\TeX}}}

\graphicspath{{figures/}} %bea
\usepackage{xcolor} %bea
\usepackage{subcaption} %bea
\usepackage{hyperref} %bea
\usepackage{rotating} %bea
\usepackage{makecell} %bea
\usepackage[ruled,vlined]{algorithm2e} %bea
\usepackage{graphicx} %bea
\usepackage[normalem]{ulem}
\usepackage{ifthen}
\usepackage{multirow}

\newboolean{final}
\setboolean{final}{false}

\definecolor{darkgreen}{rgb}{0.0, 0.5, 0.0}
\definecolor{cadded}{rgb}{0.0,0.8,0.0}
\definecolor{cdeleted}{rgb}{0.8,0.0,0.0}

\newcommand{\ric}[1]{\textcolor{orange}{(Ric) #1}}

\newcommand{\summary}[1]{}

\newenvironment{SUBENVhighlight}[1]{\color{#1}}{\color{black}}

\newcommand{\added}[1] {\ifthenelse{\boolean{final}}{~#1}{\begin{SUBENVhighlight}{cadded}~#1\end{SUBENVhighlight}}}
\newcommand{\delete}[1]{\ifthenelse{\boolean{final}}{   }{\begin{SUBENVhighlight}{cdeleted}\sout{#1}\end{SUBENVhighlight}}}
\newcommand{\replace}[2]{\ifthenelse{\boolean{final}}{#1}{\added{#1}\delete{#2}}}

\usepackage[shortcuts]{extdash}
\usepackage{paralist}

\newcommand{\anova}[5]{{\small $F_{#1,#2}\,$=$\,#3$, $p\,#4\,#5$}} % Anova "significatif" : DF, DFDen, F, p (sign,
\newcommand{\anovaRic}[5]{{\small $F_{#1,#2}\,$= $#3$, $p\,#4\,#5$}} % Anova "significatif" : DF, DFDen, F, p (sign,

\newcommand{\QF}{$QF$} %bea

\setcopyright{acmcopyright}
\acmJournal{PACMCGIT}
\acmYear{2021} \acmVolume{4} \acmNumber{3} \acmArticle{}
\acmMonth{9} \acmPrice{15.00}\acmDOI{10.1145/3480136}

\begin{document}

\title[A Perceptually-Validated Metric for Crowd Trajectory Quality Evaluation]{A Perceptually-Validated Metric for Crowd Trajectory Quality Evaluation}

\author{Beatriz Cabrero Daniel}
\email{beatriz.cabrero@upf.edu}
\orcid{0000-0001-5275-8372}
\affiliation{%
  \institution{Universitat Pompeu Fabra}
  \city{Barcelona}
  \country{Spain}
}

\author{Ricardo	Marques}
\email{ricardo.marques@ub.edu}
\orcid{0000-0001-8261-4409}
\affiliation{%
  \institution{Universitat de Barcelona}
  \city{Barcelona}
  \country{Spain}
}

\author{Ludovic Hoyet}
\email{ludovic.hoyet@inria.fr}
\orcid{0000-0002-7373-6049}
\affiliation{%
  \institution{Inria, Univ Rennes, CNRS, IRISA}
  \city{Rennes}
  \country{France}
}

\author{Julien Pettr\'e}
\email{julien.pettre@inria.fr}
\orcid{0000-0003-1812-1436}
\affiliation{%
  \institution{Inria, Univ Rennes, CNRS, IRISA}
  \city{Rennes}
  \country{France}
}

\author{Josep Blat}
\email{josep.blat@upf.edu}
\orcid{0000-0002-5308-475X}
\affiliation{%
  \institution{Universitat Pompeu Fabra}
  \city{Barcelona}
  \country{Spain}
}

\renewcommand{\shortauthors}{Cabrero, et al.}

\begin{abstract}
    Simulating crowds requires controlling a very large number of trajectories and is usually performed using crowd motion algorithms for which appropriate parameter values need to be found. The study of the relation between parametric values for simulation techniques and the quality of the resulting trajectories has been studied either through perceptual experiments or by comparison with real crowd trajectories. In this paper, we integrate both strategies. A quality metric, \QF{}, is proposed to abstract from reference data while capturing the most salient features that affect the perception of trajectory realism. \QF{} weights and combines cost functions that are based on several individual, local and global properties of trajectories. These trajectory features are selected from the literature and from interviews with experts. To validate the capacity of \QF{} to capture perceived trajectory quality, we conduct an online experiment that demonstrates the high agreement between the automatic quality score and non-expert users. To further demonstrate the usefulness of \QF{}, we use it in a data-free parameter tuning application able to tune any parametric microscopic crowd simulation model that outputs independent trajectories for characters. The learnt parameters for the tuned crowd motion model maintain the influence of the reference data which was used to weight the terms of \QF{}.
\end{abstract}




%bea
\begin{teaserfigure}
    \centering
    \includegraphics[width=0.9\linewidth]{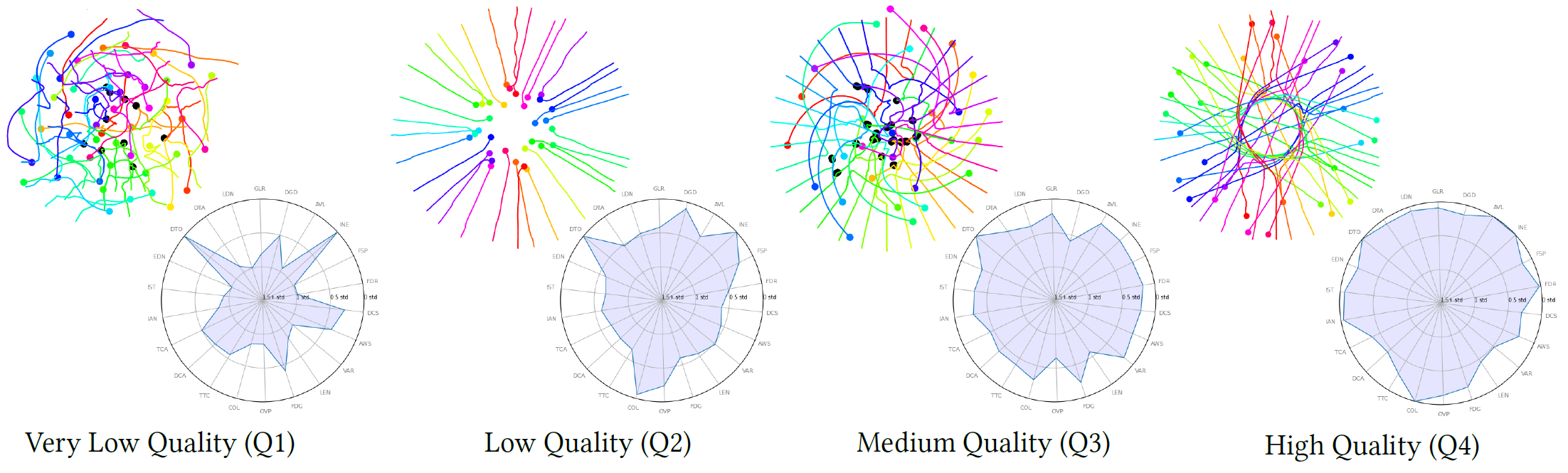}
    \caption{Crowd simulations with different quality levels. Coloured points represent characters and black points represent collisions (top). The initial and goal positions are the same in the four trajectories and simulations are of the same time-length. Radar plots (bottom) represent trajectory features values, used in the evaluation.}
    \Description{Lines of different colours representing trajectories of characters moving in a circle crossing scenario.}
    \label{fig:trajectoryRadarPlotTeaser}
\end{teaserfigure}

\maketitle

\section{Introduction}

Crowd simulators are useful to populate large environments with characters, e.g. to create lively and plausible scenes. It has long been established that the algorithms and their parameters have a direct impact on the resulting quality of animations. To assist designers in their task, we explore the issue of automatically evaluating the quality of character trajectories in a crowd, and the effect of parameters values. Two paradigms confront each other, each with its advantages and weaknesses.
Perception studies allow to directly judge the quality of animations as perceived by spectators, and therefore perfectly fulfill the objective, but are carried out in long cycles and only allow to estimate parameters that are fixed in advance. 
Methods of comparison with real data allow adjusting the parameters to each case, but data is required and the question of comparison criteria arises. Note that the comparison of data also requires objective metrics to measure how distant simulations are from real trajectories. Many metrics have been proposed, but their links with the perceived quality of animations have not been clearly established yet.

In this paper, we address the problem of evaluating the quality of crowd simulations by exploring a new method that gathers the advantages of these 2 paradigms. 
Our approach is to first establish a list of crowd trajectory features which are likely to impact the quality of crowd animations. They are established in discussion with crowd animations experts and concern individual, interaction or global crowd motion features. In a second step, we build a so-called Quality Function \QF{} which maps those features to a single quality score. This function has parameters, the value of which is established from our experts' feedback or estimated from real data. Finally, we perform a perception study with naive users so as to demonstrate that our quality function returns values that correlate with the perceived quality of crowd animations. We demonstrate our approach on the case of \emph{ambient crowds} that can be loosely described as trajectories of heterogeneous pedestrians in the street without any specific behaviour other than walking to their goal (no queuing, no running, no grouping, etc.). For this case, we learn value ranges for each term of \QF{} from real data, we establish a relationship between model parameters and perceived quality, and automatize evaluation to autonomously find parameters for models.

Our main contribution is a method to automatically evaluate crowd simulation trajectories that is in line with trajectory quality perception. We also contribute in a broader way with
(i) conclusions from a survey with crowd simulation experts about trajectory evaluation, the results of which can further guide future research on perception-driven evaluation of trajectories,
(ii) a parametric function for trajectory evaluation that can be trained with existing data sets and used to evaluate new trajectories without relying on real data directly anymore,
(iii) a user study to establish correlations between the output of the evaluation and the perception of human trajectory quality,
and (iv) a practical application relying on our quality function \QF{} to optimise the parameters of any crowd simulation model without relying on real data directly.
\section{State of the art} \label{sec:soa}

The crowd simulation research field is concerned with understanding, predicting and reproducing the motion of real human crowds. Crowd simulators are based on several classes of algorithms which are designed to generate realistic trajectories of numerous moving characters. The notion of realism is discussed below. Various approaches to this problem have been proposed. \emph{Macroscopic} approaches consider crowds as a whole, modeling it as a single continuous moving matter~\cite{Hughes2003,Treuille2006}. \emph{Microscopic} crowd simulation algorithms set the principles by which agents move individually and global crowd motion effects are expected to emerge from the interactions between agents. The seminal work of Reynold~\shortcite{Reynolds1987} explored how to control boids by following the mean velocity field generated by neighbors. The number of categories of simulation algorithms rapidly grew with force-based models \cite{Helbing1995,Karamouzas2014}, velocity-based models \cite{Paris2007,VandenBerg2008,Karamouzas2009}, vision-based models \cite{ondvrej2010synthetic,Dutra2017}, or data-driven models \cite{Lerner2007,charalambous2014pag}. These are few examples of a large body of literature. 

Crowd simulations result in large sets of individual animation trajectories. Their quality depend on a number of rules by which agents move (simulation models), as well as parameter values to control the simulation. They are not intuitive nor easy to tune and often depend on context. Our objective is to propose a method to evaluate these simulation results, regardless of the method by which they are generated. We focus on the visual realism of a crowd, i.e. the judgement from spectators of whether a simulation seems real. 
We can distinguish various approaches to the evaluation of crowd simulations. The first uses paths of real crowds, to evaluate the ability of simulators to reproduce them. The question of comparison metrics is central, and several adapted solutions have been proposed \cite{guy2012statistical,Wolinski,Charalambous2014DD}: these metrics consider crowd movement at different scales and take into account the variability of behaviors. However, there are drawbacks associated with the use of reference data, e.g., over-fitting due to the limited sample of pedestrian trajectories available. A broader perspective has been adopted by some authors that, instead of focusing on agent trajectories, measure crowd motion characteristics such as the ratio between the density and the average speed in different cultures~\cite{fundamentalDiagrams, fundamentalDiagramsCultures}. 

To overcome the problems related to the availability of crowd data, some authors study properties that trajectories should exhibit. For instance, Guy et al.~\shortcite{pledestrians} propose a metric of the effort agents expend in a path of a particular length that is based on the deviation from the shortest path to the destination and from the preferred speed. Berseth et al.~\shortcite{steerfit}, extend the effort metric together with other metrics (e.g., path length, failure rate, similarity to ground truth data) to automatically tune models and to decide what tuned motion model fits a situation better. Kapadia et al.~\shortcite{steerbug} propose objective measurable features to compare crowd simulations to real data and detect unrealistic patterns. Another strategy consists in proposing representative scenarios as a benchmark and studying the coverage of different motion models (scenarios they can handle) \cite{scenarioSpace}. Many of the measures used in these works to evaluate the performance of models are discussed and combined in this paper.

A second category of approach, rather than looking for criteria or data capable of determining the level of realism of a simulation, is to directly evaluate the perceived realism through perception. For instance, McDonnell et al.~\shortcite{mcdonnell2008, mcdonnell2009} explore the impact of character appearance and motion variations on the perception of crowd heterogeneity, while Turnwald et al.~\shortcite{Turnwald2015} propose a metric to reproduce the human perception of motion differences. The perception of human motion animation in relation with collisions, that are trajectory events, has also been studied~\cite{hoyet2016, kulpa2011}. However, to the best of the authors knowledge, there is no proven link between autonomous quality evaluation methods and human perception of trajectory quality.
Finally, one should notice the recent evolution of simulation techniques towards data-driven models. Recent approaches based on deep learning such as Social-LSTM or Social-GAN and consors~\cite{alahi2016social,gupta2018socialgan} make an implicit evaluation of the generated trajectories (through the loss function, or the discriminator component of a GAN). These methods are rather used to solve trajectory prediction problems, but their use can be adapted to the synthesis of crowd trajectories \cite{Amirian2019_GAN}. 

\paragraph{Rationale.} Our approach integrates the two first types of approaches for trajectory evaluation, and is designed to preserve the advantages of each: anchoring in objective reference data like data-driven methods, generalising with the help of experts the concept of realism through measurable trajectory features, and linking the evaluations with spectators through a perceptual study. 
With this approach, evaluations are fast to compute and provide intuitive results, which is very useful to computer animators, while still retaining the information from real data. We demonstrate its usefulness to determine simulation parameter values. 
\section {Overview} \label{sec:overview}
The objective of this work is to propose a metric to evaluate the quality of a set of 2D trajectories resulting from crowd simulation. Through the following of this paper, we call this metric the Quality Function (\QF{}), where by \emph{quality} we mean the \emph{level of perceived realism} of a given trajectory according to general users. However, since determining the relevant trajectory features for estimating such a quality is not straightforward, we propose to rely on a 2-step process. In a first step, we select relevant motion characteristics with the help of experts in the fields of Crowd Simulation and Human Animation. Then, in a second step, the selection is validated through a questionnaire that a different set of experts are asked to fill. This feature selection process, together with the features' impact on quality perception according to experts, is detailed in Section~\ref{sec:expert-analysis}. Once these expert-based relevant features are selected, we measure them in real data and propose a novel quality metric, the Quality Function \QF{}, as described in Section~\ref{sec:quality-function}. In Section~\ref{sec:validation}, through a perceptual experiment, we show that the proposed metric is able to capture the human perception of trajectory quality. Moreover, Section~\ref{sec:applications} illustrates a practical application for automatically determining the parameters of a crowd simulation model by maximising \QF{}. Finally, we present our conclusions in Section~\ref{sec:conclusion}.
\section{Trajectory features selection} \label{sec:expert-analysis}

\begin{table}
  \caption{Trajectory features discussed with experts (Section~\ref{sec:expert-analysis}) and used in the quality function (Section~\ref{sec:quality-function}), referred throughout the paper using the associated 3-letter codes.} \label{tab:feature-codes}
  \resizebox{\linewidth}{!}{\begin{tabular}{lll}
    \toprule
    \multicolumn{3}{c}{\textbf{Individual features (Code)}} \\ 
    \midrule 
    Average walking speed (AWS) & Difference to goal direction (DGD) & Inertia (INE)  \\ 
    Flickering in direction (FDR) & Flickering in speed (FSP) & Goal reaching (GLR) \\ 
    Difference to comfort speed (DCS) & Angular velocity (AVL) & Trajectory length (LEN)\\ 
    \toprule
    \multicolumn{3}{c}{\textbf{Interaction and global* features (Code)}} \\
    \midrule
    Environment-based density (EDN) & Number of collisions (COL) & Local density (LDN) \\
    Distance to other agents (DTA) &  Time to collision (TTC) & Interaction strength (IST) \\
    Time to closest approach (TCA) & Personal space overlap (OVP) & Fundamental diagram (FDG)* \\
    Interaction anticipation (IAN) & Distance at closest approach (DCA) & Feature values variety (VAR)*\\
   \bottomrule
\end{tabular}}
\end{table}

\summary{Step 1: finding features.} 
The first step in the development of the QF is to determine what crowd motion characteristics affect the perception of quality. 
\summary{From literature.} 
On the one hand, there is an important body of literature providing metrics to evaluate and compare steering algorithms, e.g., absolute difference to ground truth \cite{Wolinski}, coverage \cite{steerfit}, etc. These studies provide insights into what is desirable in a crowd motion. 
\summary{Experts helped.} 
Metrics extracted from related literature, though, are numerous and their definitions often overlap. To only keep trajectory features that have an impact on trajectory quality and are not redundant, we conducted face-to-face interviews with 6~professionals (6+ years of experience) from different leading international animation and crowd simulation companies.
\summary{Step 2: balancing the features.} 
The second step is to understand how important each characteristic is and how it can be measured in 2D reference data and synthetic trajectories. To this end, a different set of 9~experts were consulted to validate the features importance and discuss their admissible value ranges. 

\subsection{Face-to-face interviews with experts}
\label{sec:face-to-face}

To identify trajectory characteristics of crowd motions that are relevant to measure quality, 6 experts were interviewed. Prior to the interview, they received a general explanation of the project. Then, the interviewees selected a number of \emph{trajectory features} to discuss which they believed to affect perceived realism of a simulated crowd, and which can be classified into three groups (see Table~\ref{tab:feature-codes}): (i) individual trajectory features, which measure the trajectory of a single agent independently, e.g. difference between walking and comfort speed; in contrast; (ii) interaction features that deal with measures taking into account any pair of agents; the last group is for (iii) global trajectory features aggregate these measurements and study their distribution among the crowd or study the relation between one or more trajectory features. 
These features were discussed to understand their impact on the perceived quality of crowd motion. Experts were also asked about admissible value ranges for the discussed features and sufficient conditions to consider a crowd trajectory to be of low quality. Some higher-level properties such as coherence in time, the animation layer and scene decorations were also discussed with experts but are not included in this work as they were considered to be outside the scope of ``simulating ambient crowds''. 
Note that some of the proposed features are interrelated, e.g., fundamental diagrams, the relation between a crowd's flow speed and density, depends on the values of two features. This is very important for there are behaviours that are undesirable even if some features are individually inside their acceptable value range, but their interrelation in the context of ambient crowds is not.

\subsection{Questionnaire for experts}
\label{sec:expert_questionnaire}

Following the interviews, we created an online questionnaire for experts in the fields of crowd simulation and character animation, including questions about all the features that had been deemed to be important. The objective was to obtain qualitative and quantitative information about trajectory features in each group, as well as to ask for additional details, e.g. about their relative importance. 
After providing information about their background and occupation, participants then answered questions about the trajectory features listed in Table~\ref{tab:feature-codes}. Questions about each trajectory feature were grouped and displayed on the same page, beginning with relevant descriptions and video examples. The full questionnaire is provided in the Supplementary Material.

In each page, experts were asked a number of questions about the related features, providing their answer using a 7-point scale, ranging from ``totally disagree'' to ``totally agree''. Experts were asked about the features' importance under different conditions. Note that the formulation of the questions varied slightly to adapt to the feature definitions and units. The goal was to gain a more detailed understanding of how different features and feature values would impact the perceived quality. Moreover, a couple of open answer questions were asked in each page of the questionnaire for experts to give their general opinion about the proposed trajectory features, including their opinion about values, related features, etc. 
After the feature-specific questions, the experts then answered questions about the general appearance of the crowd, e.g. heterogeneity, as well as to the relative importance of trajectory features. Questions about the impact of trajectory features and of specific values of the features were answered using the following scale: ``not important at all'', ``of very little importance'', ``of little importance'', ``of average importance'', ``important'', ``very important'', and  ``absolutely essential''. A different set of experts (9) participated in the survey. Table~\ref{tab:expertBackgroundField} reports their expertise, showing that most respondents worked in the industry (6).

\begin{table}
    \centering
    \caption{Reported expertise (from 1 to 5) on Human Animation and Crowd Simulation of the 9 participants.} 
    \label{tab:expertBackgroundField}
    \begin{tabular}{lcc}
    \toprule
      \textbf{Participant Field}& \multicolumn{2}{c}{\textbf{Reported expertise in}} \\
      \textbf{of Expertise} & \textbf{Human Animation} & \textbf{Crowd Simulation} \\
      \midrule
        Crowd Simulation Research (3) & \{4, 4, 5\} & \{4, 5, 5\} \\
        Video-Game Industry (3) & \{2, 4, 5\} & \{2, 3, 4\} \\
        Movie Industry (3) & \{3, 4, 4\} & \{2, 4, 5\} \\
        \bottomrule
    \end{tabular}
\end{table}

\begin{figure*}[t]
    %\centering
    %    \begin{subfigure}[b]{\linewidth}
    \centering
    \includegraphics[width=\linewidth]{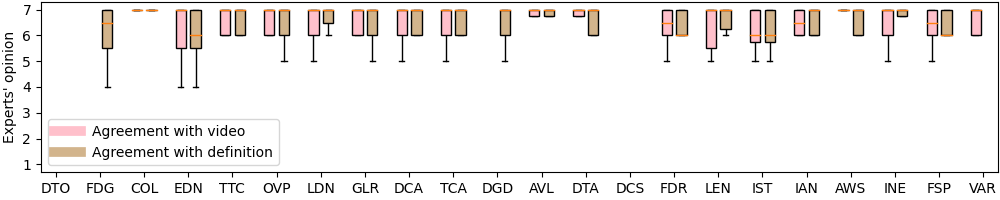}
    \hfill
    \includegraphics[width=\linewidth]{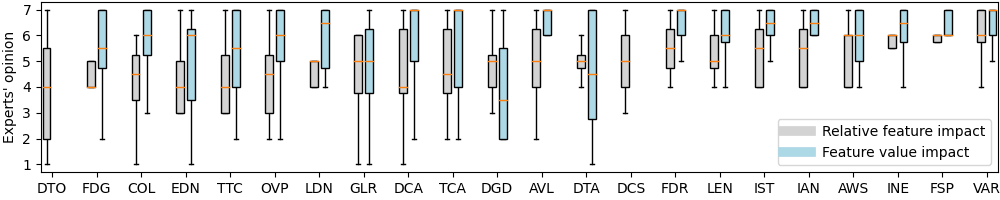}
    \caption{Agreement of experts with the statements about feature definitions (top, brown), video examples (top, pink), values (bottom, blue), and importance (bottom, grey). Labels in the horizontal axis correspond to the 3-letter codes in Table~\ref{tab:feature-codes}. Boxes represents the central 50\% of the answers and whiskers, the remaining 50\%. Orange bars represents the median value. The absence of a box means that there was no question in the survey about the definition, video, or importance of that feature.}
    \label{fig:expertSurvey}
\end{figure*}

\subsection{Discussion of expert opinion and of feature selection}

Figure~\ref{fig:expertSurvey} (top) shows on the vertical axis the agreement of experts on a scale from 1 to 7 when questioned about each feature on the horizontal axis. It can be, therefore, said that participants in the survey agreed with the definition of all the selected trajectory features (definitions in Supplementary Material), and that the example videos were representative of the trajectory features in question. This means that the concept of features used for trajectory evaluation was understood and that the mathematical definition of the features is in line with the behaviours and artifacts shown in the videos. Moreover, the survey also validated the use of \textit{ChAOS}\footnote{https://project.inria.fr/crowdscience/project/ocsr/chaos/} to showcase desired undesired trajectory feature values in human trajectories. 

Figure~\ref{fig:expertSurvey} (bottom) illustrates the experts' opinion on the importance of features. Experts were also asked whether particular values for the trajectory features affected the perceived quality. A minority of experts indicated that some trajectory features were ``of little'' or ``of very little'' importance for perceived quality of trajectories. However, all features were on average considered of being at least ``of average importance'' (values over 4 on the vertical axis), and some ``very important''. Their opinion on feature value impact (blue) suggests that values outside the admissible range for any of the selected features negatively affect the perceived quality. With this, we conclude that all the features contribute to some degree to the perceptual quality of trajectories, and should therefore be used in the evaluation of crowd trajectory quality.

% ANSWERS TO OPEN QUESTIONS AND OPINIONS ABOUT VALUE RANGES AND MEASUREMENTS
Answers to the open questions also provided us with valuable information.
% TTC 
For instance, E1 highlighted the importance of the available animation system in handling collisions. In his point of view, collisions are ``not particularly important if the animations are appropriate for both characters.'' E2 also reported on this effect by reminding that it is common to ``keep agents that intersect slightly because our eye assumes they are not colliding, (as) the human eye often does not register collisions.'' Such comments are in agreement with previous works on this topic~\cite{hoyet2016} and suggest a difference between intersections (personal space overlap) and actual perceivable collisions (collisions), and as reported by E4 ``collisions do happen in real life, so some tolerance is required''. However, when such animations are not available, or if the system is not dealing with collisions, ``the collision then should be avoided all together, and having a high TTC value would be better'' (E1). In this paper we focus on trajectories, without making assumptions about the animation system used to visualise characters on such trajectories. Together, these comments highlight the importance of preventing collisions when evaluating trajectories. 

% ANGULAR VELOCITY
However, the importance of avoiding collisions might be mitigated by other factors, as E2 also emphasized the importance of avoiding sharp changes in direction (i.e., high angular velocity values), by commenting that ``it is better to collide than to have an abrupt change of direction.'' E1 also mentioned that it is preferable to avoid sharp changes in velocity, unless ``we have captured animations for this and can play those animations appropriately.'' Similarly, E3 reported that ``animation plays a big part'' when a collision is imminent (small TTC) and that ``sidestepping is perceived as more realistic than turning completely 90 degrees to avoid the collision.'' %
% INTERACTION ANTICIPATION
Another expert, E5, also commented that ``abrupt turns are not found in real trajectories'', and that ``there is a maximum turning angle to avoid obstacles, the primary reaction is always slowing down.'' In his point of view ``humans have an anticipation of about 30 meters from their surroundings'', which could provide an estimate of the distance up to which an agent can be affected by its neighbours. 

% TRAVELLING TIME
Experts also commented on the actual trajectory lengths and directions to reach a goal. In particular, according to E2 ``limiting travel time to a specific duration makes the resulting simulation look artificial'', for it would penalise trajectories going around obstacles and favouring the shortest path and higher walking speeds, which is not necessarily close to that of a real human. E4 also mentioned that ``unnecessarily large desired direction differences might work just as well.'' These comments suggest that trajectory lengths need to be varied enough to avoid impairing realism, and not moving towards the goal at all times might be preferable in some contexts. 
% WALKING SPEED
Of course, travel time is also affected by the walking speed of characters, which experts agreed should be within an admissible range. In particular, they voiced their concerns about displaying walking characters moving very slowly or very fast, as ``changing the speed of the animation does affect visual quality'' (E1) and ``animations cannot be re-timed to change the speed because of motion dynamics'' (E2). 

% DENSITY
Interestingly, experts were unanimous about environment\-/based density: the distribution of inter-pedestrian distances has an impact on realism since an uniformly distributed crowd is not believable. Many participants voiced their concerns about distances among groups of pedestrians as well. E2 said that people might move in groups and that, depending on the context, ``local density among grouping agents might be high, and the closest distance very small.'' Another expert (E4) suggested that density should be computed as ``the ratio of flesh per square meter to account for children'', while also highlighting that there are many types of pedestrians for which different state feature values are expected such as ``children, disabled people, inebriated people, or people queueing'' (E5 adds ``menacing people'' and ``elderlies'' to the list). Similarly, E4 pointed out that ``fundamental diagram is also a cultural trait'', highlighting that cultural differences might also play a role in evaluating trajectory features. Such comments highlight the importance of individual differences, even though we focus (as a first step) on ambient crowds of similar-sized adults without any specific behaviour or cultural traits.

\section{Quality Function} \label{sec:quality-function}

% INTRO TO GOAL: MEASUING TRAJECTORY QUALITY
Our goal is to propose a function for an autonomous quantitative evaluation of crowd trajectory quality. In this context, every character in a crowd has a trajectory in the studied time window, i.e., the length of the reference video or the simulation length, and the term \emph{character trajectory} is used here to refer to the sequence of 2D positions of one character together with information about its state at every time step. The term \emph{crowd trajectory} is then used to refer to the set of all character trajectories of a crowd.
% LINK TO PREVIOUS SECTION QUALITY FUNCTION BUILT ON TRAJECTORY FEATURES PICKED BY EXPERTS
%\summary{Function built on the trajectory features.} 
The work presented in Section~\ref{sec:expert-analysis} provides us with a set of trajectory features, the relevance of which for assessing the believability of a \emph{crowd trajectory} has been validated by experts. These are the basic building blocks with which to construct our quality function. However, to effectively build it, the trajectory features need to be measured, ranked and combined to create the \emph{quality function}. 

% PHASES OF THE EVALUATION PROCESS
Constructing \QF{} is a three-fold process. First, metrics for quantitative evaluation for each of the trajectory features are proposed, and real data is studied to obtain reference values for the features. These reference values are treated as a golden set, used to inform the quality function about typical and expected values. With this information, a cost to penalise deviations between the golden set and the features of the evaluated trajectory is proposed. Finally, these independent feature costs are combined through a weighted sum to determine the quality of the evaluated trajectory.

\subsection{QF definition} % FORMULAS FOR QUALITY FUNCTION > INDEPENDENT COST FUNCTIONS
% THE QUALITY IS A SUM OF COSTS
The quality function, \QF{}, is defined as one minus a weighted sum of costs ($C_i$), where the subscript $i$ indicates the index of the trajectory feature. $C_i$ depends on the distribution of feature values in ground truth data, $r_i$, and in the trajectory being evaluated, $s_i$. \QF{} is therefore defined as: 
\begin{equation} \label{eq:quality-function}
    QF = 1 - \sum_i{\omega_i C_i(s_i|r_i)},
\end{equation}
where $i \in [1, \ldots, I]$ is the trajectory feature index, $I$ is the number of features ($I$=21 in our examples), and $\omega_i$ is the weight associated with the feature $i$. The $\omega_i$ values are automatically learnt as discussed in the following sections.
% TRAJECTORY FEATURE COSTS % TRAJECTORY FEATURE COSTS % TRAJECTORY FEATURE COSTS 
\summary{Two types of costs.} As identified in Table~\ref{tab:feature-codes}, trajectory features can be classified into three categories: individual, interaction and global. Individual and interaction features take different values for each character and time-step of the trajectory while global features take different values only for each time-step of the trajectory. \summary{Time and character dependent costs.} For those trajectory features with values changing for each character, $n$, and time-step, $t$, we compute the cost as follows:
\begin{equation} \label{eq:cost-function-normal}
    C_i(s_i|r_i) = \sum_{n} {\sum_{t} { \left( 1 - e^{ -{(s_i^{n,t}-\mu_i)^2 } / {2\sigma_i^2} } \right)}} / (NT),
\end{equation}
where $N$ is the number of characters in the crowd, $T$ is the length of the simulated trajectory, and $\mu_i$ and $\sigma_i$ are the average and standard deviation of the trajectory feature values found in real data $r_i$. \summary{Global feature costs.} For trajectory features computed across characters at each time-step, $t$, we compute:
\begin{equation}
    C_i(s_i|r_i) = \frac{1}{T}\sum_{t} { \left( 1 - e^{ -{(s_i^{t}-\mu_i)^2 } / {2\sigma_i^2} } \right) }.
\end{equation}
\summary{This is not arbitrary. Justification needed? Appendices?}
\summary{Get feature values before computing $C_i$.} The trajectory feature values for $s_i$ and $r_i$ are determined prior to evaluating trajectories using \QF{}. For this, we assume future linear motion of all the characters in the crowd. We move on to discuss the feature values and the distributions found in reference trajectories.

% STUDYING TRAJECTORY FEATURE VALUES AND REFERENCE DATA 
% STUDYING TRAJECTORY FEATURE VALUES AND REFERENCE DATA 
% STUDYING TRAJECTORY FEATURE VALUES AND REFERENCE DATA 

\subsection{Trajectory feature values} \label{sec:feature-values}

The values of the features in the evaluated trajectory, $s_i$, are obtained using the state properties of all the characters. Moreover, the trajectory feature values are compared in the cost functions $C_i$ to a gold distribution, $r_i$, which we compute in practice from a set of reference data. The dataset used in this work is pending publication; meanwhile, it is available on demand. Following Eq.~\eqref{eq:cost-function-normal}, we are interested in the average value, $\mu_i$, and standard deviation, $\sigma_i$, for each of the trajectory features in real trajectories. 

%\summary{Types trajectory properties / states.} REMOVED COMMENT FROM PREVIOUS VERSION/ORDER, THANKS
Character properties can be classified in: static, individual and state properties. Properties like agent radius and mass are static: unchanging and shared by all agents. Individual properties, such as goal position or preferred speed, are different for each character in the crowd but remain constant throughout the trajectory. Lastly, state properties refer to the properties of a character that change at every time-step of a trajectory, e.g. current speed, direction, position, etc. Static, individual and state motion properties are used to compute the values of \emph{trajectory features}. 
Each feature value set $s_i$ will be the result of measuring the values of a particular trajectory feature, $f_i$, at each time-step, $t$, and for each character in the crowd, $n$, if appropriate:

\begin{equation}
    s_i=\{f_i(n,t)\}_{n\in[0\cdots N],t} % j is for each character, t for timestep, q = "function of the state"
    % Julien's version: s_i=\{ q_{j,t}\}_{(j,t)}.
\end{equation}
\summary{The time window for $s_f$ depends on the application.} Note that the values of $s_i$ change at every time step as characters move and interact. The time window is not specified in Eq.~\eqref{eq:cost-function-normal} and greatly depends on the application as will be discussed in Section~\ref{sec:validation}.
A detailed account of how the values for each trajectory feature are computed is given by the mathematical definitions provided in Supplementary Material.

\begin{figure}
    \newcommand\sizeplots{0.35\linewidth}
    \centering 
    \begin{subfigure}[b]{\sizeplots}
        \centering
        \includegraphics[width=\linewidth]{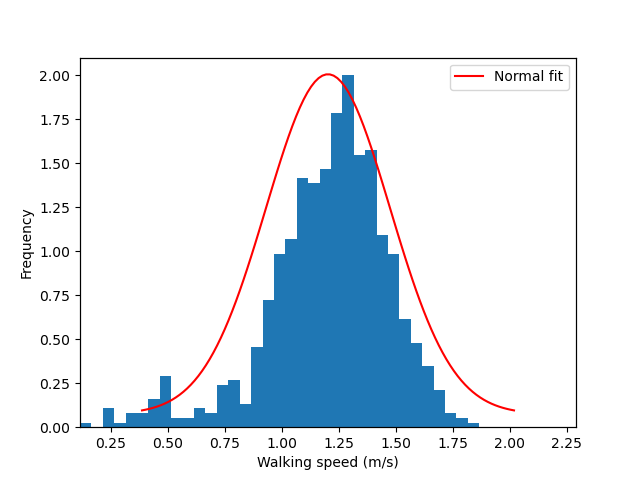}

        \caption{Real data (reference).}
        \label{fig:histogramAndFit:left}
    \end{subfigure}
    \hspace{1em}
    \begin{subfigure}[b]{\sizeplots}
        \centering
        \includegraphics[width=\linewidth]{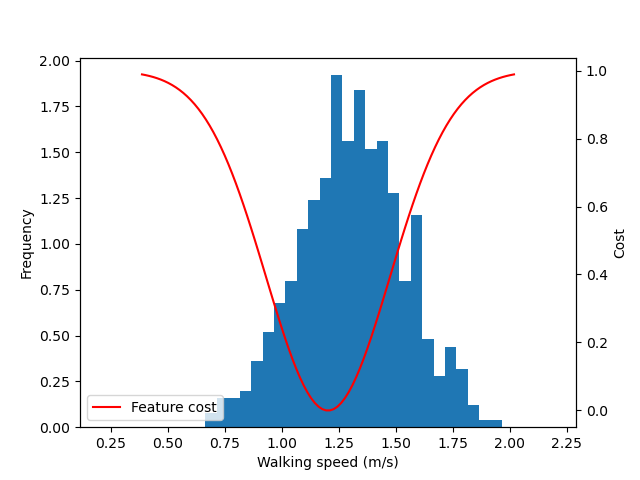}

        \caption{Synthetic data.}
        \label{fig:histogramAndFit:right}
    \end{subfigure}
    \caption{Blue: walking speed distribution in reference data (a) and in a synthetic trajectory (b). Red: normal distribution fitted to the reference data (a), and the trajectory cost assigned to each value (b).}
    \label{fig:histogramAndFit}
\end{figure}

Walking speed is an important term of the quality metric and is used here to illustrate the analysis of feature value distribution. The velocity of the characters of the crowd, the output of the motion model, is observable at every time-step of the simulation, and we can therefore use the values of the walking speed for all characters for all the time-steps of the trajectory. Figure~\ref{fig:histogramAndFit:left} shows the distribution of walking speed values in reference data (blue bars), as well as the normal distribution fitted to this data (red line); the corresponding mean ($\mu_{ws}$) and standard deviation ($\sigma_{ws}$) of this distribution will be used to compute the cost for walking speed $C(s_{ws}|r_{ws})$. Figure~\ref{fig:histogramAndFit:right} shows the distribution of walking speed values found in a sampled synthetic crowd trajectory (blue bars), as well as the cost assigned to each feature value according to Eq.~\eqref{eq:cost-function-normal} (red curve). Intuitively, the larger the difference between the feature distribution in the simulation and in the reference data, the higher the cost in Eq.~\eqref{eq:cost-function-normal}.

Analysis of the feature values in reference data $r_i$ revealed distributions close to that of a normal distribution (example in Figure~\ref{fig:histogramAndFit}). In some cases the feature values exhibit a multi-modal shape which would be better explained with a Gaussian mixture, but upon inspection these trajectories were identified to be outside the scope of this project (ambient crowds).

\subsection{Quality function weights} \label{sec:learning-weights} % TRAINING THE QUALITY FUNCTION
\summary{Need to learn weights of QF.} The weight $\omega_i$ of each cost in Eq.~\eqref{eq:quality-function} is related to the influence of each trajectory feature on the quality of crowd trajectories. These weights are learnt using an evolutionary approach that leads to a single set of weights used in \QF{}. 
Different learning strategies can be used to give values to the \QF{} weights, but Genetic Algorithms (GA) are not prone to be stuck in local minima and provided stable results in our experiments. GA are able to explore new weight value sets and exploit the solutions found in previous iterations through mutation and crossover strategies.

\summary{Fitness measurement.} In this work, the performance of characters is measured on a given training set. The training set consists of two types of trajectories: (i) reference data, that receive a score equal to 1, and (ii) unrealistic trajectories, according to experts, that are given a low score. 
For instance, trajectories generated without avoidance maneuvers are studied and the corresponding feature values are used as predictors and assigned 0 as target value in the training set. Moreover, undesired feature value combinations (discussed in Section~\ref{sec:expert-analysis}) are also given a 0 target score. In this step we use different scenarios: crossing flows, circle crossing and random. The consistency of the learnt weights across scenarios is discussed in Section~\ref{sec:validation}.

The average difference between the given quality score, $S_{R}$, and the prediction, $S_{QF}$, is used as the fitness of each individual of the population. For information, the weights learnt for each trajectory feature are reported in Table~\ref{tab:cost_weights}; where we see that some features have a more direct impact on the overall evaluation. Nevertheless, all features contribute to the quality score in a significant way, allowing it to detect artifacts that simpler metrics such as the Least Effort Function~\cite{pledestrians} could not e.g. unrealistic combinations of speed and density or the apparition of flickering in some interactions. Some examples of these artifacts and their compared evaluation are discussed in the Supplementary Materials.
Correlations between features were also studied prior to learning the weights, to avoid highly correlated features to impair learning, but that none had a correlation coefficient greater than 0.8. All the features were therefore included to be potential predictor variables. 
We would like to point out that the Distance To Obstacles (DTO) feature was selected by experts (Table~\ref{tab:feature-codes}) but was not used in our experiments because relevant information was not available in our reference dataset (our reference trajectories did not include obstacles other than neighbouring characters).
Nevertheless, weights for such features can be learned in the future if a different dataset including the relevant information is made available.

\begin{table}
  \caption{Weights for Eq.~\eqref{eq:cost-function-normal} costs, that depend on the feature values in synthetic and reference data.}
  \label{tab:cost_weights}
    \begin{tabular}{l|lr|lr|lr|lr|lr}
        \toprule
        \multicolumn{11}{c}{\textbf{Trajectory feature weights}}\\
        \midrule
        Individual & AWS & 0.1995 & DCS & 0.0258 & FDR & 0.0590 & DGD & 0.0072 & GLR & 0.0074\\
        & FSP & 0.1054 & INE & 0.0275 & AVL & 0.0800 & LDN & 0.0381 \\
        \midrule
        Local & DTA & 0.0586 & EDN & 0.0087 & IST & 0.0163 & TTC & 0.0441 & COL & 0.0949\\
        & IAN & 0.0085 & TCA & 0.0698 & DCA & 0.0068 & OVP & 0.0096 \\
        \midrule
        Global & FDG & 0.0515 & LEN & 0.0587 & VAR & 0.0224 & & \\ 
        \bottomrule
    \end{tabular}
\end{table}

\begin{comment}
\begin{table*}
  \caption{\ric{Alternative version.} Weights for the trajectory feature costs described in Eq.~\eqref{eq:cost-function-normal} that in turn depend on the distribution of the feature values found in synthetic and reference data.}
  \label{tab:cost_weights}
    \begin{tabular}{lr|lr|lr|lr|lr|lr|lr}
        \toprule
        \multicolumn{14}{c}{\textbf{Trajectory feature weights}}\\
        \midrule
        SPW & 0.1995 & SPC & 0.0258 & FDR & 0.0590 & FSP & 0.1054 & INE & 0.0275 & AVL & 0.0800 & DDD & 0.0072 \\
        GLR & 0.0074 & DNL & 0.0381 & DTA & 0.0586 & DNE & 0.0087 & IST & 0.0163 & IAN & 0.0085 & TCA & 0.0698 \\
        DCA & 0.0068 & TTC & 0.0441 & COL & 0.0949 & OVP & 0.0096 & FDG & 0.0515 & LEN & 0.0587 & VAR & 0.0224 \\ 
        \bottomrule
    \end{tabular}
\end{table*}
\end{comment}

\subsection{Consistency check with real data}

Once the weights are learnt, crowd trajectories are given a quality score using Eq.~\eqref{eq:quality-function}. The quality score obtained in real trajectories is studied using cross-validation. This test confirms that previously unseen real trajectories receive scores higher than synthetic data, $S_{QF} \sim \mathcal{N}(0.9160,\, 0.0817)$. 
Whilst this alone is not enough to confirm the correctness of the quality function, it partially substantiates its usefulness. The following parts of this paper, Section~\ref{sec:validation} and Section~\ref{sec:applications}, describe in greater detail the validation and applications of the proposed quality function.

\section{Perceptual validation} \label{sec:validation}

To the best of the authors knowledge, none of the existing data-driven metrics for trajectory quality is proven to accurately characterize human perception of trajectories. In the previous sections, a new quality metric \QF{} was proposed, its weights were learnt using reference data and used to assign scores to new trajectories. In this section, we are interested in whether the obtained score correlates with human evaluations of trajectory quality. To answer this question, we conducted an online experiment, where participants were shown pairs of videos and asked to select the more realistic one.

\subsection{Experiment design}

Sixty-six participants took part in our perceptual experiment. All participants (26F/39M/1N; age: 30$\pm$9, min=16, max=61) were non-experts in crowd simulation or related fields, and were naive to the purpose of the study. Participants were only informed at the beginning of the experiment that it was about trajectory realism and that data would be treated anonymously. The experiment was conducted through an ad hoc responsive website and the replies were collected on a private server.

\summary{Visualized using ChAOS. } The experiment consisted in showing to participants a number of pairs of videos, presented side by side. The focus of this work is the ``realism'' of trajectories, not of the animation of characters. To help participants focus on the trajectories, animations therefore had to be believable without masking the underlying trajectory artifacts. Following recommendations on variety in crowds~\cite{mcdonnell2009}, trajectories were animated with a set of 20 male and 20 female characters, with 6 textural variations per character. 

\begin{figure}
    \newcommand\sizeScenario{0.25\linewidth}
    \centering
    \includegraphics[width=\sizeScenario]{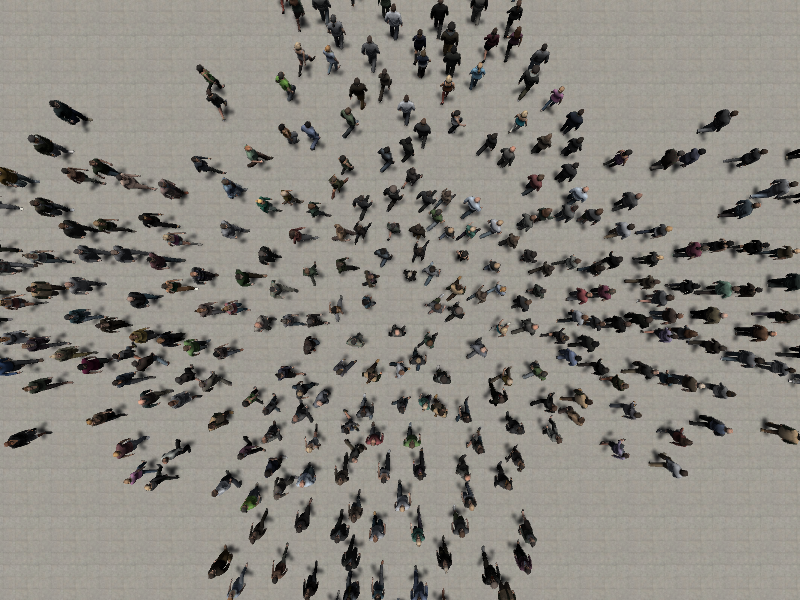} \hfill
    \includegraphics[width=\sizeScenario]{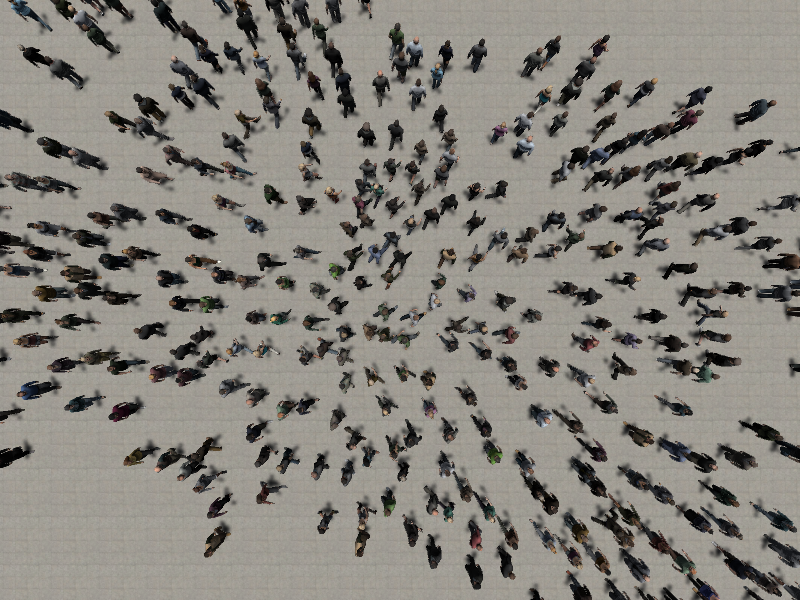} \hfill
    \includegraphics[width=\sizeScenario]{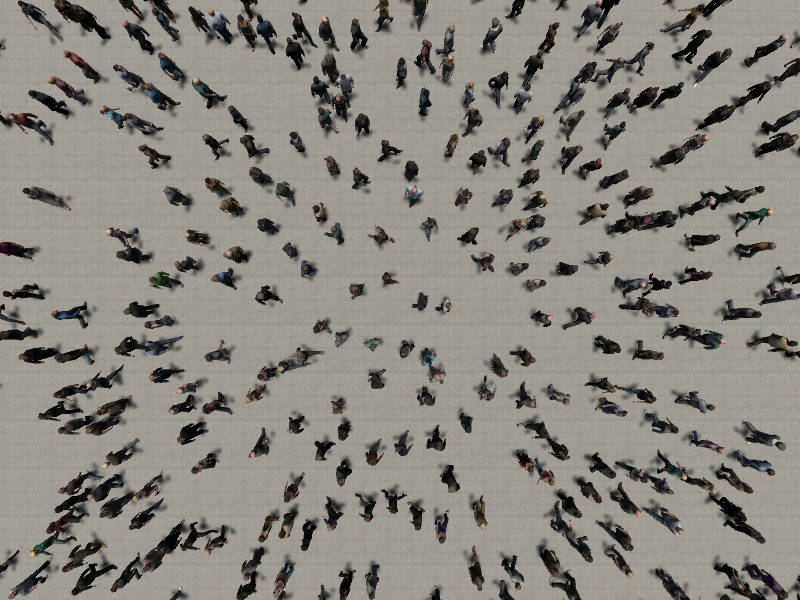} \\
    \includegraphics[width=\sizeScenario]{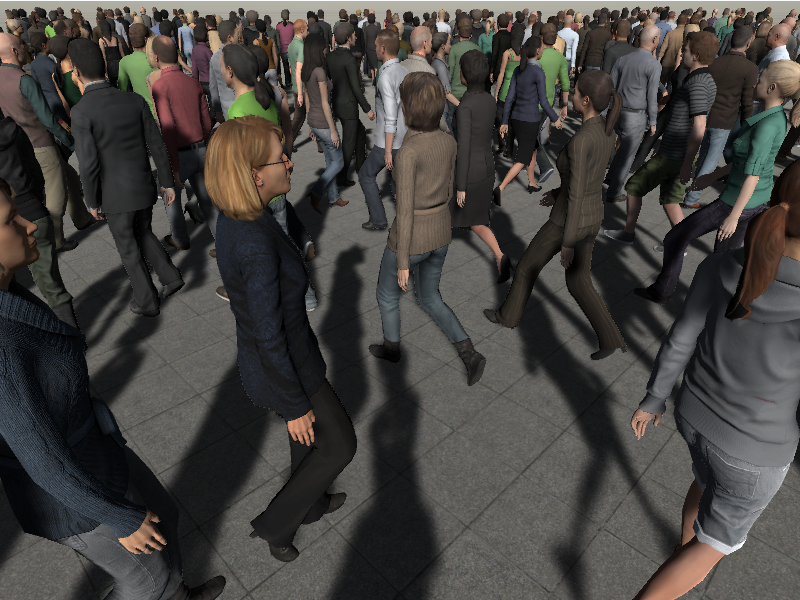} \hfill
    \includegraphics[width=\sizeScenario]{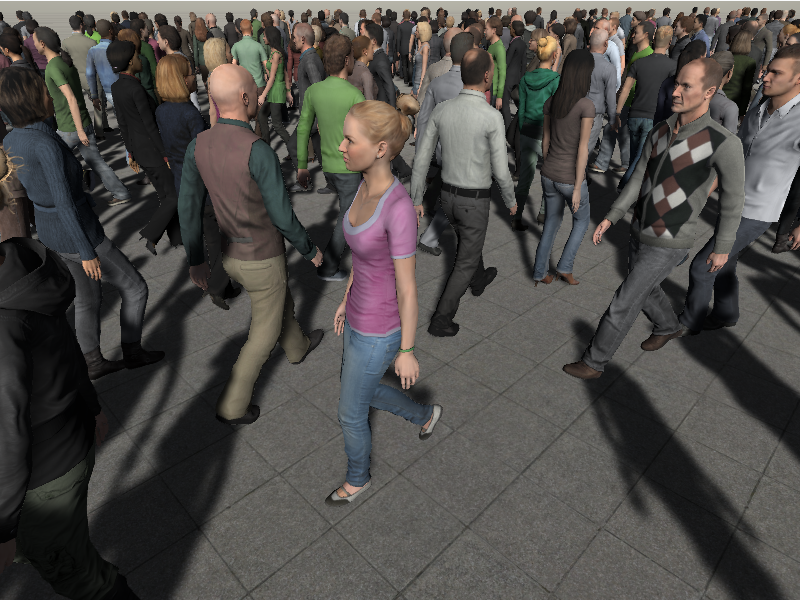} \hfill
    \includegraphics[width=\sizeScenario]{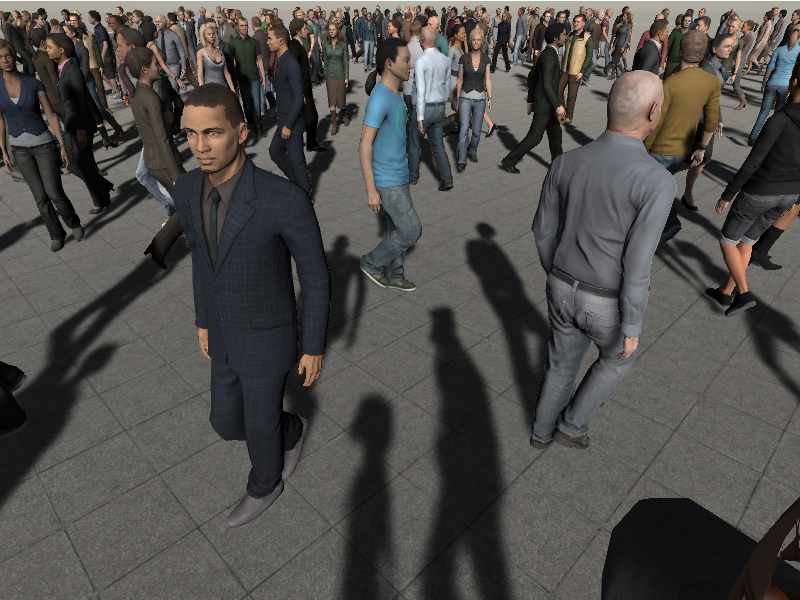}
    \caption{Scenarios and point of views used in our experiment. From left to right: 2-flows 90º crossing, 2-flows 135º crossing and random scenarios, displayed from top (top row) and eye level (bottom row) point of views.}
    \label{fig:naiveQuestonnaire}
\end{figure}

There are factors that can influence the perception of quality. For instance, different motion models generate different types of trajectories, with different motion characteristics. Therefore, in order to make the experiment robust, we selected three representative models to generate the experiment videos: (i) a representative of force-based models, Social Forces; (ii) a representative of velocity-based approaches, RVO; (iii) and a linear prediction-based model, Dutra-Marques. Density can also impact the perception of artifacts so crowds were generated at two density levels: high (more than 3p/$m ^2$) and low (less than 1p/$m^2$). To study a wider range of interactions, three generic scenarios were also selected: a random one, and two crossing flows at 90º and 135º, respectively (see Figure \ref{fig:naiveQuestonnaire}). Furthermore, previous work has focused on determining the influence of the point of view on the perception of realism; to ensure the camera position did not affect the results, we therefore chose two points of view (see Figure~\ref{fig:naiveQuestonnaire}): a canonical eye-level camera and a top view. 

Trajectories used in this experiment obtain \QF{} scores in the [0, 0.9) range. To study the perception of trajectory visual quality, the videos were classified into quartiles (Q1=lowest quality quartile, Q4=highest quality quartile). As we were interested in evaluating perceptual differences between the quartiles, we included in our experiment comparisons between all quartiles (6 combinations). All combinations between these factors were taken into account to create the stimuli, resulting in 288 videos (5 seconds, $960\times1080$ pixels) created using \textit{ChAOS}: 4 quality levels $\times$ 3 scenarios $\times$ 3 motion models $\times$ 2 densities $\times$ 2 POV $\times$ 2 repetitions.

The experiment was divided into three blocks (one for each scenario). Each participant performed a single block of the experiment, and was shown a random subset of 72 video pairs for this specific scenario (a total of 144 good and bad quality trajectories). Each pair of videos showed crowd motions with different quality scores Q$i$ and Q$j$, but generated with the same motion model, scenario, density level, and point of view. In the following, such a comparison is referred to as Q$i$Q$j$. The left/right position of the videos on the screen was randomly assigned for each trial. There was no time limit for completion of the experiment, and the critical time of this experiment, i.e. the minimum time needed for users to watch each pair at least once, is 12 minutes. After watching each video in a pair, participants were asked ``Which one looks more realistic to you?'', and answered by clicking with the mouse on a button to select a video of the pair. The mouse position was reset to the center of the screen between each trial by pressing a 'Next' button, to avoid any left/right selection bias.

Once a participant completed the experiment, a table was stored on the server-side containing, for each trial, the ids of the videos shown, the participant answer, as well as the side of the screen the selected video was shown on (left/right). For each participant, the time for experiment completion was also stored, together with the age and gender, asked at the end of the experiment. Replies were then aggregated to study correlations between the participant choices and the quality function score. Out of 66 participants, 24 performed the random scenario, 22 performed the 90º crossing flows scenario, and 20 performed the 135º crossing flows scenario.

\subsection{Statistical analysis} \label{sec:userexp-stats}

\summary{Accuracy definition.} To analyse participants' results, we first determine True Positive (TP) answers, i.e., participant selects the video with the highest \QF{} score. With this information, \textbf{user accuracy} is computed as the percentage of TP for each user, and used to evaluate the level of agreement of users with \QF{}.

To check for statistical differences in user accuracy, we performed on each within-subject factor (Q$i$Q$j$ Comparisons, Motion Models, Densities, POV) a separate 2-way mixed-design repeated measures Analysis of Variance (ANOVA) with between-subject factor Scenario. A full factorial analysis was however not possible as participants were presented with a random subset of 72 video-pair comparisons. 
Normality was assessed using a Kolmogorov-Smirnov test. All effects were reported at $p$<0.05. When we found effects, we further explored the cause of these effects using Bonferroni post-hoc tests for pairwise comparisons. As we did not find any interaction effect of Scenario and Motion Model on user accuracy, these are omitted in the following of the section. 

~\\\textbf{Q$i$Q$j$ Comparisons.} We first looked at statistical differences between the Q$i$Q$j$ comparisons, to evaluate effects of quality comparisons on user accuracy. Results showed a main effect of Q$i$Q$j$  (\anova{5}{300}{10.551}{\approx}{0}). Actual accuracy values per Q$i$Q$j$  comparisons are reported in Table~\ref{tab:qualityPreference}. Post-hoc analysis showed that all Q$i$Q$j$ comparisons were perceived on average with a similar accuracy, except for Q1Q2 for which participants demonstrated a significantly lower accuracy. This result suggests that participants had more difficulties in differentiating between videos of the Q1 and Q2 quartiles, but that they were more accurate for comparisons between other quartiles, reaching a 75\% accuracy for comparisons between Q1 and Q4. Single t-tests on accuracy were also significantly higher than 50\% for each Q$i$Q$j$ comparison (all $p<0.05$), showing that user accuracy was in all cases significantly above chance level. 

To better understand the relationship between user preference and quality, we averaged accuracy over trials with the same quartile difference. We organised the data in three Quality Difference levels: QD1 (over Q1Q2, Q2Q3, Q3Q4), QD2 (over Q1Q3, Q2Q4) and QD3 (Q1Q4). A repeated measures Analysis of Variance (ANOVA) with within-subject factor QD showed a main effect of QD on user accuracy (\anova{2}{120}{25.253}{\approx}{0}). As quality differences are directly related to our \QF{} scores, this result demonstrates that users were more likely to correctly identify the highest \QF{} quality videos when the \QF{} quality difference between videos increased (Figure~\ref{fig:QQMainEffect}), and therefore to agree with the \QF{} scores. 

\begin{table}
\begin{minipage}{0.45\linewidth}
\centering
    \includegraphics[width=\linewidth]{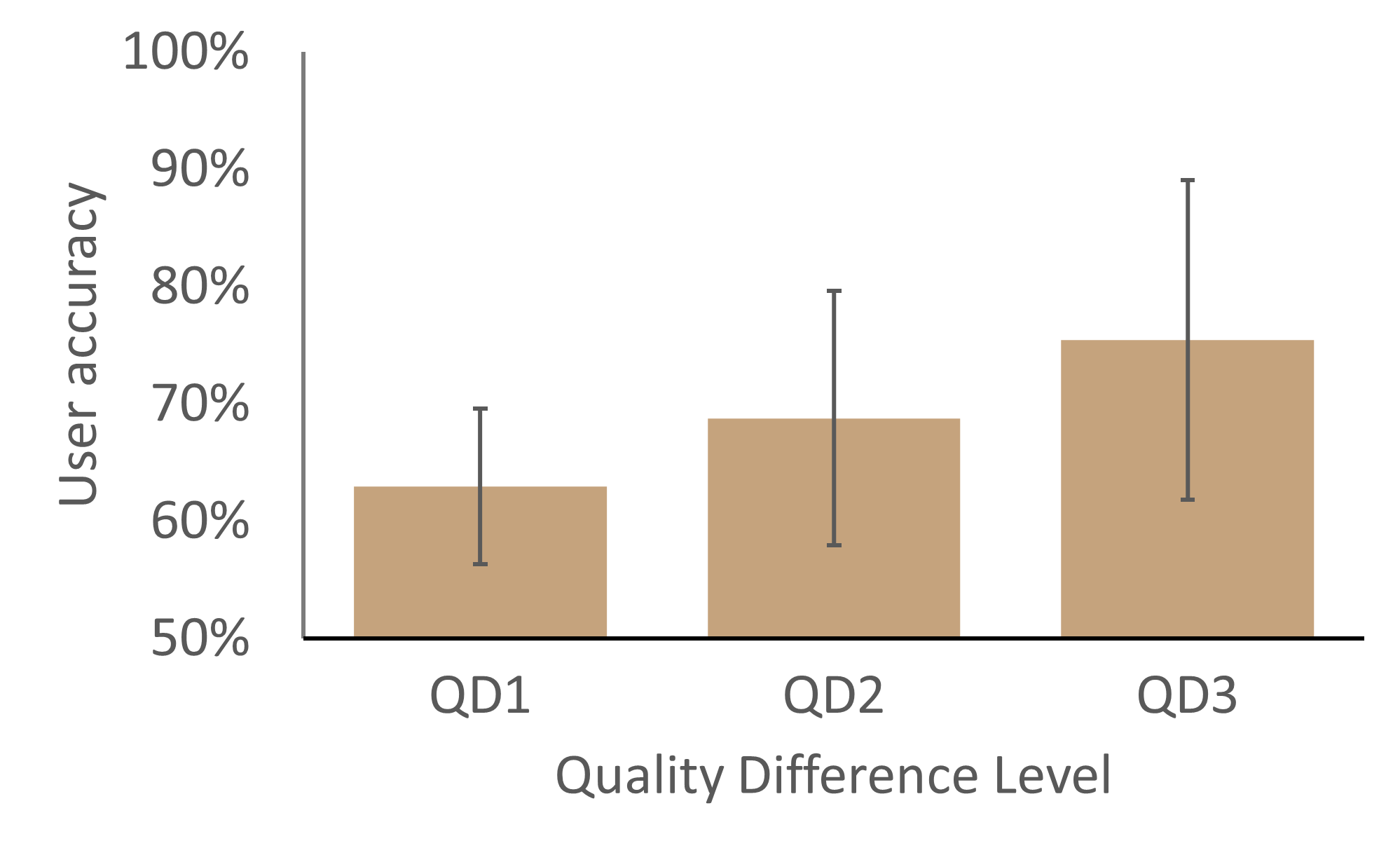}
     \captionof{figure}{Effect of quality difference on user accuracy. Error bars depict the standard deviation. }
     \label{fig:QQMainEffect}
\end{minipage}\hfill
\begin{minipage}{0.5\linewidth}
\centering
    \caption{User accuracy detailed per quartile-to-quartile comparisons.}
	\label{tab:qualityPreference}
	\begin{tabular}{cc|cccc}
        \toprule
        & & \textbf{Q1} & \textbf{Q2} & \textbf{Q3} & \textbf{Q4} \\
        \midrule
        \parbox[t]{2mm}{\multirow{4}{*}{\rotatebox[origin=c]{90}{Preferred}}} 
        & \textbf{Q1} & & & & \\ 
        & \textbf{Q2} & 55$\pm$20\% & & & \\ 
        & \textbf{Q3} & 68$\pm$16\% & 63$\pm$14\% & &  \\ 
        & \textbf{Q4} & 75$\pm$15\% & 68$\pm$15\% & 67$\pm$17\% & \\ 
        \bottomrule
    \end{tabular}
\end{minipage}
\end{table}

~\\\textbf{POV \& Density.} Interestingly, we found a main effect of POV (\anovaRic{1}{60}{27.892} {\approx}{0}), showing that user accuracy was on average higher for the eye-level (72$\pm$7\%) than for the top (64$\pm$8\%) point of view. We also found a main effect of Density (\anova{1}{60}{12.996}{\approx}{0}), showing that user accuracy was on average higher when presented with higher than lower density video pairs (64$\pm$9\% vs. 70$\pm$8\%). We did not find a significant effect of any of the other factors on the perception of trajectory quality.

\subsection{Discussion}

The results of the user evaluation demonstrated that viewers consider, on average, videos with a higher \QF{} score to be more realistic than videos with a lower \QF{} score, therefore showing their general agreement with the quality function. This agreement can be observed in the Q$i$Q$j$ comparisons (Figure~\ref{fig:QQMainEffect}, left), where accuracy was significantly above chance level in all cases, and up to 75\% accuracy on the higher quality differences. We also observed that accuracy significantly increased depending on the difference in \QF{} scores between the presented videos. Overall, these results validate that the output of our data-driven \QF{} and the perception of human trajectory quality are in agreement.

However, the user evaluation also demonstrated that other factors can influence the perception of the visual realism of trajectories. Such factors include elements that are not captured by \QF{} and are depending on application choices, such as the point of view. For instance, an eye-level point of view seems to enable users to more accurately perceive visual artifacts in character trajectories, while a top-down viewpoint might render perceiving such artifacts more difficult. We believe this is because users are able to assess the quality of trajectories from a more ``natural'' point of view, closer to the one we are used to experience as humans. The top view, on the other hand, somewhat masks the artifacts in the trajectories, making it more difficult to evaluate the crowd motion. Such results are therefore complementary to the work of Kulpa et al. ~\shortcite{kulpa2011}, who showed that collisions are more easily spotted by viewers from a top point of view. 
Nevertheless, other factors also seem to influence the perception of trajectory features. For instance, the results show that users more easily perceived quality differences for higher density scenarios. Such a difference could be due to several possibilities, such as higher numbers of collisions in higher density scenarios, possibly leading to more sudden direction and velocity changes in low quality trajectories, etc., and should therefore be explored in future studies.
Similarly, the range of tested scenarios did not allow us to evaluate the effect of all the trajectory features, since the study was limited to ambient crowds. For instance, the goal to reach was not displayed, which is however a feature influencing $S_{QF}$ score. It is thus unknown whether or to what extend such features affect the perception of trajectory quality.

% APPLICATION % APPLICATION % APPLICATION 
% APPLICATION % APPLICATION % APPLICATION 
% APPLICATION % APPLICATION % APPLICATION 

\section{Practical Application} \label{sec:applications}
In this section, we illustrate one of the possible applications of the proposed metric: tuning parameters of crowd motion models without direct comparisons to real trajectories, i.e., using the pre-trained \QF{}. The goal is therefore to find a parameter set, $p^{opt}$, of a crowd motion model that maximises the \QF{} score for a generated trajectory or for a set of trajectories. Resorting to \QF{} (which can be seen as a fit-function) instead of directly fitting the model to real trajectories has several advantages. 
First, in this work, the information in the reference data is transformed into abstract trajectory features selected by experts. This makes the learning process focus only on properties that affect the perception of crowd motion quality instead of simply trying to replicate real trajectories which are, in fact, difficult to gather. 
Second, over-fitting is a recurrent problem when learning crowd motion model parameters. Models usually describe a limited set of interactions and learning the parameters to replicate real data usually leads to ad hoc solutions. In this work, we propose an abstraction from real data so that the algorithm does not replicate actual trajectories, but instead mimics the values of several abstract properties.

\subsection{Parameter tuning}
\QF{} depends on a number of crowd property values which are initialized using real data (see Section~\ref{sec:learning-weights}). Theoretically, parameters for any microscopic crowd motion model can be learnt using \QF{}, by evaluating the feature values in the simulated trajectories. The learning algorithm would then attempt to maximise \QF{} by changing the model parameters. 
In this work, we propose a learning process with a decreasing exploration rate to find parameters, in order to explore a wide parameter domain while still ensuring convergence. However, other global optimization techniques can be applied to maximise \QF{} for any trajectory. Finding the most adequate method is, in itself, an interesting topic but out of the scope of this paper. 
It is important to note that, when using different parameter value sets for each agent in the crowd, the complexity of the optimization explodes due to the large number of parameters to learn, because of the typically large number of characters.

\subsection{Learning process}
There are two possible goals of the learning process: (i) finding $p^{opt}$ for a particular initialization in a given scenario; (ii) finding a generic $p^{opt}$ which can be used for any ambient crowd scenario. In the former goal, a single initialisation (specifying a scenario, a simulation algorithm and the internal characteristics of characters, e.g., initial position, comfort speed, goal direction) is used for tuning the model using \QF{}. In this case, the resulting $p^{opt}$ might not be useful with another crowd initialisation. In the latter goal, the objective is to find a scenario-independent $p^{opt}$ for a selected simulation algorithm, instead of focusing on a particular initialization. During training, multiple crowds are simulated at every iteration to evaluate the performance of each parameter set. The evolutionary process ends when the stop criteria are met. In this case, the genetic algorithm stops iterating if it has reached a user-defined maximum number of iterations, if the fitness increase is below a threshold for several iterations, or if the quality reaches the maximum score.

\subsection{Results} 
Figure~\ref{fig:trajectoryRadarPlotTeaser} shows an example of the learning process at different stages using the Dutra-Marques model in a circular scenario (characters initially positioned on a circle and assigned the goal of reaching the diametrically opposite position). In this example, characters share parameter values.
Four steps of the learning process are shown (one per column), with increasing quality scores from left to right (each column illustrates a quality quartile Q1-Q4, as in Section~\ref{sec:userexp-stats}). The resulting trajectories, all of the same time length, are shown in the top row, while radar-plots showing the difference to ideal feature values are shown below. 
At the beginning of the training, the pool of parameters is unlikely to contain parameter values leading to good quality simulations. In Figure~\ref{fig:trajectoryRadarPlotTeaser} (top, left), parameters related to goal attraction are not good to guide agents towards their goal, on the other side of the circle. The corresponding radar plot shows that the trajectory features are very different to those found in real data which results in a low \QF{} score for this trajectory ($S_{QF}=0.22$). \QF{} is specially affected by the number of local features like overlaps and interaction strength (see Table~\ref{tab:cost_weights}).
As training progresses and the pool of parameters evolves, the resulting simulations improve with respect to \QF{} and characters progressively learning to balance between avoiding collisions, attempting to reach their goal, and satisfying the other conditions, e.g., acceptable walking speed, jerkiness. As a result, the right-most plot ($S_{QF}=0.79$) shows a trajectory created with parameters that balance these implicit rules. More examples of this process are in the Supplementary Material.

\subsection{Discussion}

Once the weights are established, \QF{} can be used to train crowd simulation models \emph{without further requiring any real data}. This is one of the main advantages of our approach, as it facilitates parameter tuning for crowd simulation models. Consequently, training the crowd simulation parameters is, in our approach, data-free, while they are strongly influenced by the real data features captured by \QF{}. 
However, training a crowd simulation model using \QF{} also has limitations. The main one is the ability of a simulation model to express a large variety of agent interactions. The learning algorithm might output parameter values leading to low $S_{QF}$ values if it is unable to improve the quality score for a maximum number of iterations. For example, in Figure~\ref{fig:trajectoryRadarPlotTeaser}(bottom, right-most), some trajectory feature values differ from those found in reference data, e.g., average walking speed (faster) and Distance at Closest Approach (smaller). However, in the corresponding trajectories (top row), we see that the learnt parameters led to collision-free, smooth trajectories. This means that the genetic algorithm was able to find a parameter set $p^{opt}$ to maximise \QF{} but, due to the limitations of the model, the parameter sharing, and the symmetry of the scenario, the quality could not be further improved. 
Finally, the learning results can also be influenced by the scenarios used in the training phase. To cope with this, we use multiple scenarios generated with random initial positions and goals.
\section{Conclusion} \label{sec:conclusion}

We presented a new perceptually-validated quality metric to automatically evaluate crowd trajectories. The proposed evaluation is designed with the help of experts and abstracts from real data. A number of trajectory features, deemed relevant to capture perception of trajectory quality, were selected with the support of experts. Then, measurements were proposed to learn the typical feature values in real data. These features were then combined in the so-called quality function, \QF{}, which returns a score from 0 to 1 for any 2D ambient crowd trajectory. To validate \QF{} we conducted a web-based experiment with non-expert users and we show that there is a high agreement between viewers' perceptions of visual quality and \QF{} scores. Finally, we demonstrate a practical application for tuning parameters of crowd simulation models using~\QF{}.

Real trajectory data is only required when training \QF{}. Once this is done, \QF{} can be independently used to evaluate new trajectories, leveraging the feature information previously captured using real data. Besides, our feature-based approach allows avoiding over-fitting the parameters to a specific scenario or dataset by not using reference data directly. 
With this paper, we provide a fully-functional, pre-trained, \QF{} which is ready to use by the scientific community. Note however that the provided set of feature weights can be easily re-targeted to accommodate other use cases. Examples of these use cases could be to include reactions to obstacles, or specific behaviours such as queuing, which were not available in our training dataset. It it also important to point out that specific feature value ranges and weights can be manually adjusted while training \QF{}, which opens the door to automatic generation of, e.g., non-realistic animations requiring overly-smooth trajectories, or simulating a marathon requiring specific speed ranges.

Finally, \QF{} has been proved to capture the main characteristics which affect human perception of trajectory quality. Nevertheless, a more complex combination of features could be explored to be able to evaluate other scenarios and behaviours (e.g., different types of agents, grouping, queuing), even though the combination of the features might be application dependent. 
As an alternative to tuning motion models directly, the more intuitive \QF{} can be manually tuned by artists to fit a particular scene, such as for further penalizing collisions, or enforcing goal reaching.

We have discussed how \QF{} is useful for ambient crowds, which do not cover all possible crowd compositions or behaviours. Moreover, there is a significant body of work that has shown that not all pedestrians use the same steering strategies in all scenarios (it depends on factors such as age, abilities, cultural differences, etc.). In order to deal with diversity in complex scenarios, future research could focus on a \QF{}-based policy adaption strategy dependent on character properties and their environment. For instance, this could enable characters in low density areas to use simpler motion models, while characters in higher density areas would use more complex models. In order to do this, it might be interesting to model the feature costs as a sum of Gaussians, instead. Another approach could be to derive a crowd motion model that, in order to compute the optimal velocity at each time step, maximises \QF{}.

\begin{acks}
With partial support of the EU funded project PRESENT, H2020-ICT-2018-3-856879. As Serra Húnter Fellow, Ricardo Marques acknowledges the support of the Serra Húnter Programme to this work.
\end{acks}

\bibliographystyle{ACM-Reference-Format}
\bibliography{sample-base}

\end{document}